%% file: paper.tex
\documentclass[]{damo}

\usepackage{wrapfig}
\usepackage{tabularx}
\usepackage{textcomp}
\usepackage{stfloats}
\usepackage{url}
\usepackage{verbatim}
\usepackage{graphicx}
\usepackage{titlesec}
\usepackage{tocloft}
\usepackage{adjustbox}
\usepackage{multirow}
\usepackage{pifont}
\usepackage{tikz}
\usepackage{comment}
\usepackage{amsmath,amssymb} 
\usepackage{colortbl}  
\usepackage{color}
\usepackage{booktabs} 
\usepackage{hyperref}
\usepackage{graphicx}    
\usepackage{subcaption} 
\usepackage{booktabs}
\usepackage{amsmath} 
\usepackage{multirow} 
\usepackage{booktabs} 
\usepackage{subcaption} 
\usepackage{amsthm}

\definecolor{diffcolor}{RGB}{255, 224, 224} 

\definecolor{codeblue}{rgb}{0.25,0.5,0.5}
\definecolor{codekw}{rgb}{0.85, 0.18, 0.50}

\definecolor{codesign}{RGB}{0, 0, 255}
\definecolor{codefunc}{rgb}{0.73, 0.33, 0.83}
\lstdefinelanguage{PythonFuncColor}{
  language=Python,
  keywordstyle=\color{blue}\bfseries,
  commentstyle=\color{codeblue},  
  stringstyle=\color{orange},
  showstringspaces=false,
  basicstyle=\ttfamily\small,
  literate=
    {*}{{\color{codesign}* }}{1}
    {-}{{\color{codesign}- }}{1}
    {+}{{\color{codesign}+ }}{1}
    {dataloader}{{\color{codefunc}dataloader}}{1}
    {sample_two}{{\color{codefunc}\texttt{sample\_two\_time}}}{1}
    {sample_one}{{\color{codefunc}\texttt{sample\_one\_time}}}{1}
    {randn}{{\color{codefunc}randn}}{1}
    {randn_like}{{\color{codefunc}randn\_like}}{1}
    {jvp}{{\color{codefunc}jvp}}{1}
    {stopgrad}{{\color{codefunc}stopgrad}}{1}
    {metric}{{\color{codefunc}metric}}{1}
}

\RequirePackage{xspace}
\makeatletter
\DeclareRobustCommand\onedot{\futurelet\@let@token\@onedot}
\def\@onedot{\ifx\@let@token.\else.\null\fi\xspace}

\def\eg{\emph{e.g}\onedot}

\makeatother

\definecolor{adptorange}{RGB}{248, 205, 172}
\definecolor{cmpblue}{RGB}{189, 215, 238}
\definecolor{cmpblue}{RGB}{189, 215, 238}

\definecolor{our_red}{RGB}{232,157,160}
\definecolor{our_blue}{RGB}{136,206,230}
\definecolor{our_orange}{RGB}{246,200,168}
\definecolor{our_green}{RGB}{178,211,164}

\definecolor{attn_code0}{RGB}{247,215,200}
\definecolor{attn_code1}{RGB}{238,169,139}
\definecolor{mlp_code0}{RGB}{204,201,221}
\definecolor{mlp_code1}{RGB}{102,95,153}

\definecolor{token_blue}{RGB}{84, 120, 140}

\usepackage{pifont}       
\usepackage{bbding}       
\usepackage{fontawesome}
\usepackage{xspace}

\usepackage{float}

\newlength\savewidth

\newcolumntype{x}[1]{>{\centering\arraybackslash}p{#1pt}}
\newcolumntype{y}[1]{>{\raggedright\arraybackslash}p{#1pt}}
\newcolumntype{z}[1]{>{\raggedleft\arraybackslash}p{#1pt}}

\renewcommand{\paragraph}[1]{\vspace{1mm}\noindent\textbf{#1}}
\usepackage{colortbl}
\usepackage{xcolor}
\usepackage{wrapfig}

\renewcommand{\paragraph}[1]{\vspace{1.25mm}\noindent\textbf{#1}}

\usepackage{algorithm}
\usepackage{algpseudocode}
\usepackage{listings}

\definecolor{codeblue}{rgb}{0.25, 0.5, 0.5}
\definecolor{codekw}{rgb}{0.35, 0.35, 0.75}
\lstdefinestyle{Pytorch}{
    language = Python,
    backgroundcolor = \color{white},
    basicstyle = \fontsize{9pt}{8pt}\selectfont\ttfamily\bfseries,
    columns = fullflexible,
    aboveskip=1pt,
    belowskip=1pt,
    breaklines = true,
    captionpos = b,
    commentstyle = \color{codeblue},
    keywordstyle = \color{codekw},
    mathescape = true,
}
\definecolor{green}{HTML}{009000}
\definecolor{red}{HTML}{ea4335}

\title{Few-Step Distillation for Text-to-Image Generation: A Practical Guide}

\author[* 1, 2]{Yifan Pu}
\author[* 2]{Yizeng Han}
\author[* 2]{Zhiwei Tang}
\author[\dagger 2, 3]{Jiasheng Tang}
\author[2]{Fan Wang}
\author[2, 4]{Bohan Zhuang}
\author[1]{Gao Huang}

\affiliation[1]{Tsinghua University}
\affiliation[2]{DAMO Academy, Alibaba Group }
\affiliation[3]{Hupan Lab}
\affiliation[4]{Zhejiang University \\}
\contribution[*]{Equal contribution}
\contribution[\dagger]{Corresponding author}

\input{sec/e0}

\date{\today}

\begin{document}
\thispagestyle{firstheader}
\maketitle
\pagestyle{empty}

\input{sec/e1}
\input{sec/e2}
\input{sec/e3}

\input{sec/e4}
\input{sec/e5}

\bibliographystyle{assets/plainnat}
\bibliography{paper}

\newpage
\beginappendix
\input{sec/app}

\end{document}

%% file: sec/e0.tex
\abstract{
Diffusion–distillation has dramatically accelerated class-conditional image synthesis, but its applicability to open-ended text-to-image (T2I) generation is still unclear. We present the first systematic study that adapts and compares state-of-the-art distillation techniques on a strong T2I teacher model, \textsc{FLUX.1-lite}. By casting existing methods into a unified framework, we identify the key obstacles that arise when moving from discrete class labels to free-form language prompts. Beyond a thorough methodological analysis, we offer practical guidelines on input scaling, network architecture, and hyperparameters, accompanied by an open-source implementation and pretrained student models. Our findings establish a solid foundation for deploying fast, high-fidelity, and resource-efficient diffusion generators in real-world T2I applications. Code is available on \url{github.com/alibaba-damo-academy/T2I-Distill}.
}

%% file: sec/e1.tex
\section{Introduction}

In recent years, large-scale Diffusion Models (DMs)~\citep{ho2020denoising,sohl2015deep} have achieved unprecedented success in the field of text-to-image synthesis, with generation quality that rivals or even surpasses human creation ~\citep{rombach2022high, saharia2022photorealistic}. Models such as Flux~\citep{flux1kontext2025}, Qwen-Image~\citep{wu2025qwen} and Imagen~\citep{gdm2025imagen4}, trained on massive image-text datasets, can generate high-fidelity, high-resolution images from complex textual descriptions. However, this remarkable performance comes at a significant computational cost. These models rely on an iterative sampling process, progressively converting Gaussian noise into a clear image through hundreds of Network Function Evaluations (NFEs) ~\citep{song2020denoising, ho2020denoising}. This process is not only computationally intensive but also suffers from high latency, severely hindering its application in scenarios requiring real-time feedback or operating in resource-constrained environments, such as interactive design tools, dynamic game content generation, and augmented reality. Therefore, drastically reducing generation latency while maintaining high-quality output has become a critical challenge in the field of generative AI.

To address the aforementioned challenges, a new research direction focused on few-step generation techniques has emerged~\citep{luhman2021knowledge,wang2022diffusion,salimans2022progressive}. The core objective is to develop models capable of producing high-quality images in a very small number of NFEs, typically between 1 and 8. This research direction has spurred a variety of innovative methods that, through different technical pathways, attempt to break the trade-off between quality and speed inherent in traditional diffusion models. These methods not only aim to accelerate the inference process of existing models but also explore the fundamental theory of generative models, laying the groundwork for the next generation of efficient generators.

Current mainstream few-step generation methods can be broadly categorized into three major paradigms. This report will conduct an in-depth analysis of one or more representative methods from each paradigm:

\textbf{Distribution Distillation}: This paradigm leverages a few-step and efficient student model to match the output distribution a powerful but slow pre-trained teacher model. The student model aims to simulate the teacher's output distribution in a minimal number of steps. There are two representative methods in this area:
\begin{itemize}
    \item \textbf{Direct Distribution Distillation: Distribution Matching Distillation (DMD)} is a method that distills a multi-step diffusion model into an efficient, one-step image generator by minimizing the distribution difference and incorporating a regression loss, while its successor, DMDv2, enhances image quality, training efficiency, and flexibility by replacing the costly regression loss with techniques like a GAN loss. Recent text-to-image model, like Qwen-Image-Lightning~\citep{qwenimagelig}, adopted this technique.
    \item \textbf{Adversarial Distribution Distillation: Latent Adversarial Diffusion Distillation (LADD)} distills a slow, pre-trained latent diffusion model (the teacher model) into a rapid student model capable of generating high-quality images in just one to four steps. The core of this technique~\citep{sauer2024fast} is an adversarial game within the latent space, where the student model is trained to produce outputs that fool a discriminator, forcing it to match the teacher's latent distribution and thereby achieving a massive acceleration in generation speed while maintaining high fidelity. Models like SD3-Turbo~\citep{sauer2024fast} and Flux.1 Kontext [dev]~\citep{flux1kontext2025} use this technique.
\end{itemize}

\textbf{Trajectory-based Distillation}: Trajectory distillation is a method for accelerating diffusion models by training a student model to predict an entire segment of the teacher model's sampling trajectory in fewer steps. Unlike techniques that only match the final output, this approach distills the complete generation process, which allows the student to more faithfully learn complex dynamics like classifier-free guidance and produce high-quality results in very few steps. This report will analyze the latest breakthrough in this area:
    \begin{itemize}
        \item \textbf{sCM}. Simplified Consistency Models introduces a simplified theoretical framework and a set of practical techniques to stabilize and scale the training of continuous-time consistency models, enabling them to achieve state-of-the-art performance with as few as two sampling steps in ImageNet.
        \item \textbf{IMM}. Inductive Moment Matching enforces consistency at the distribution level by using moment matching to ensure that samples generated from different points along a stochastic interpolant path converge to the same target distribution.
        \item \textbf{MeanFlow}. Mean-Flow learns the average velocity of a flow field through a novel identity, enabling stable training and achieving state-of-the-art single-function-evaluation performance.
    \end{itemize}

Distillation methods like DMD and LADD have been successfully applied to text-to-image synthesis tasks with significant results. However, the publicly available results for three highly promising new methods—sCM, MeanFlow, and IMM—are primarily focused on smaller-scale, unconditional image generation tasks (e.g., ImageNet). Their performance, adaptability, and potential advantages in the complex, open-domain task of text-to-image synthesis remain unclear. Therefore, this report aims to achieve three main goals:
\begin{enumerate}
    \item \textbf{Provide a deep theoretical comparison} of these cutting-edge methods, dissecting their fundamental differences in core mechanisms, sources of stability, and scalability.
    \item \textbf{Propose a detailed, unified, and feasible experimental plan} for adapting these methods (especially sCM and MeanFlow) to the text-to-image synthesis domain for rigorous and fair empirical evaluation.
    \item \textbf{Deliver a well-engineered, modular, and reproducible codebase} that implements the proposed adaptations, training recipes, evaluation pipelines, and baseline models, facilitating fair comparison and enabling the community to reproduce and extend our experiments.
\end{enumerate}

%% file: sec/e2.tex
\section{A brief overview of each method}
This chapter will provide a detailed analysis of the three core methods. For each method, we will explain its core mechanism, key innovations, and propose a specific, actionable plan to adapt it to the text-to-image synthesis task. This analysis is not just theoretical but aims to provide a solid foundation for the subsequent experimental design.

\subsection{stabilized Continuous-time Consistency Models (sCM)}

Consistency Models (CMs) aim to learn a function $f(\boldsymbol{x}_t, t)$ that can map a sample $\boldsymbol{x}_t$ at any noise level $t$ directly to its corresponding clean sample $\boldsymbol{x}_0$ in a single step \cite{song2023consistency}. They can be learned through distillation (Consistency Distillation, CD) or trained from scratch (Consistency Training, CT). However, early CMs, especially in the continuous-time setting, were long plagued by severe training instability \cite{luo2023lcm}. sCM introduces a comprehensive suite of techniques, including the TrigFlow formulation, architectural improvements, and an adaptively weighted training objective, to resolve the core training instabilities of continuous-time consistency models (CMs).
    

\subsection{MeanFlow}

MeanFlow takes a distinctly different approach from standard Flow Matching. Standard flow matching models aim to learn the \textit{instantaneous velocity} $\boldsymbol{v}(\boldsymbol{x}_t, t)$ of particles moving along a flow field. MeanFlow, instead, chooses to model a different physical quantity: the \textit{average velocity} between two time steps $t$ and $r$.

Beyond standard MeanFlow Training (as shown in ~\cref{alg:meanflowc2i} ~\cite{geng2025mean}), we further provide a MeanFlow Distillation algorithm. In this algorithm, we use a flow matching pretrained teacher model to provide the instantaneous velocity in the MeanFlow target. The detailed algorithm is shown in ~\cref{alg:meanflowt2i}.
    
\begin{figure}[H]
\centering

\begin{minipage}{0.47\textwidth}
    \begin{algorithm}[H]
    \caption{{MeanFlow}: Training.\label{alg:meanflowc2i}}
    \begin{lstlisting}[language=python]
# fn(z, r, t): function to predict u
# 
# x: training batch
t, r = sample_t_r()
e = randn_like(x)

z = (1 - t) * x + t * e
v = e - x

u, dudt = jvp(fn, (z, r, t), (v, 0, 1))

u_tgt = v - (t - r) * dudt
error = u - stopgrad(u_tgt)
loss = metric(error)
    \end{lstlisting}
    \end{algorithm}
\end{minipage}
\hfill
\begin{minipage}{0.47\textwidth}
    \begin{algorithm}[H]
    \caption{{MeanFlow}: Distillation.\label{alg:meanflowt2i}}
    \begin{lstlisting}[language=python]
# fn(z, r, t): function to predict u
# gn(z, t): flow matching teacher
# x: training batch
t, r = sample_t_r()
e = randn_like(x)

z = (1 - t) * x + t * e
v_teacher = gn(z, t)

u, dudt = jvp(fn, (z, r, t), (v, 0, 1))

u_tgt = v_teacher - (t - r) * dudt
error = u - stopgrad(u_tgt)
loss = metric(error)
    \end{lstlisting}
    \end{algorithm}
\end{minipage}

\end{figure}

\subsection{Inductive Moment Matching (IMM)}

IMM is a from-scratch training paradigm. Its core idea is to train the model so that the distribution of samples mapped to a target time $s$ is consistent regardless of whether they start from time $r$ or $t$ (with $s < r < t$). This inductive consistency along the time axis ensures convergence of the learned mappings. Instead of KL divergence, IMM uses Maximum Mean Discrepancy (MMD) as a stable, sample-based measure of distributional difference, which implicitly aligns all moments between distributions.

%% file: sec/e3.tex
\section{Methodological Relationships and Comparative Insights}


\subsection{FM v.s. MeanFlow: Flow Matching as a Special Case of MeanFlow (when $r \equiv t$)}

Flow Matching (Algorithm \ref{alg:flowmatching}) can be understood as a specific instance of the more general MeanFlow framework (Algorithm \ref{alg:meanflow}). 
In the general MeanFlow formulation, the model learns a average velocity vector field that depends on both the current time $t$ and a reference time $r$.
Flow Matching is precisely recovered when we set the reference time $r$ equal to the current time $t$. In this specialization, the second term of the target vanishes entirely. This simplifies the MeanFlow target to $\boldsymbol{u}_{\text{tgt}} = boldsymbol{v}$.
Consequently, the network is trained to predict instantaneous velocity directly. This is exactly the objective of standard Flow Matching, as shown in Algorithm \ref{alg:flowmatching}. A direct comparison of the pseudocode in Algorithm \ref{alg:meanflow} and Algorithm \ref{alg:flowmatching} illustrates this simplification.

\begin{figure}[H]
\centering

\begin{minipage}{0.47\textwidth}
    \begin{algorithm}[H]
    \caption{{MeanFlow}: Training.\label{alg:meanflow}}
    \begin{lstlisting}[language=python]
# fn(z, r, t): function to predict u
# x: training batch
t, r = sample_two()
e = randn_like(x)

z = (1 - t) * x + t * e
v = e - x

u_pred, dudt = jvp(fn, (z, r, t), (v, 0, 1))

u_tgt = v - (t - r) * dudt
error = u_pred - stopgrad(u_tgt)
loss = metric(error)
    \end{lstlisting}
    \end{algorithm}
\end{minipage}
\hfill
\begin{minipage}{0.47\textwidth}
    \begin{algorithm}[H]
    \caption{{Flow Matching}: Training.\label{alg:flowmatching}}
    \begin{lstlisting}[language=python]
# fn(z, t): function to predict v
# x: training batch
t = sample_one()
e = randn_like(x)

z = (1 - t) * x + t * e
v = e - x

v_pred = fn(z, t)

v_tgt = v
error = v_pred - v_tgt
loss = metric(error)
    \end{lstlisting}
    \end{algorithm}
\end{minipage}

\end{figure}

\subsection{FM v.s. sCM: Flow Matching and TrigFlow are interconvertible without re-training}

\newcommand{\bx}{\boldsymbol{x}}
\newcommand{\bz}{\boldsymbol{z}}

Flow Matching and TrigFlow frameworks are mutually convertible at inference time~\citep{zheng2025large,chen2025sana}, enabling models trained in one paradigm to be used with samplers from the other without any retraining. Specifically, a pre-trained Flow Matching model, denoted by its velocity field estimator $\boldsymbol{v_{\theta}}(\bx_{t,\texttt{FM}}, t_{\texttt{FM}}, \boldsymbol{y})$, can be used to denoise a sample $\bx_{t,\texttt{Trig}}$ from a TrigFlow process. This is achieved by first transforming the TrigFlow state $(\bx_{t,\texttt{Trig}}, t_{\texttt{Trig}})$ into its Flow Matching equivalent $(\bx_{t,\texttt{FM}}, t_{\texttt{FM}})$. The time variable is mapped to preserve the signal-to-noise ratio (SNR), and the state variable is rescaled accordingly:
\begin{equation}
t_{\texttt{FM}} = \frac{\sin(t_{\texttt{Trig}})}{\cos(t_{\texttt{Trig}}) + \sin(t_{\texttt{Trig}})}, \qquad
\bx_{t,\texttt{FM}} = \frac{1}{\cos(t_{\texttt{Trig}}) + \sin(t_{\texttt{Trig}})} \cdot \bx_{t,\texttt{Trig}}.
\end{equation}
Subsequently, the output of the Flow Matching model is used to construct the optimal TrigFlow estimator $\hat{\boldsymbol{F_{\theta}}}$, which provides the correct update direction for a TrigFlow-based solver. This relationship is given by:

\begin{equation}
\boldsymbol{\hat{F}_{\theta}}\left(\bx_{t,\texttt{Trig}},t_{\texttt{Trig}},\boldsymbol{y}\right)
= \frac{\cos(t_{\texttt{Trig}}) - \sin(t_{\texttt{Trig}})}{\cos(t_{\texttt{Trig}}) + \sin(t_{\texttt{Trig}})} \cdot \bx_{t,\texttt{FM}} + \frac{1}{\cos(t_{\texttt{Trig}}) + \sin(t_{\texttt{Trig}})} \cdot \boldsymbol{v_{\theta}}(\bx_{t,\texttt{FM}}, t_{\texttt{FM}}, \boldsymbol{y}).
\end{equation}

Conversely, this mapping is fully and losslessly reversible, allowing a native TrigFlow model $\boldsymbol{F_{\theta}}$ to operate within a Flow Matching sampler. Given a Flow Matching state $(\bx_{t,\texttt{FM}}, t_{\texttt{FM}})$, we first map it to the TrigFlow domain using the inverse transformations for time and state:
\begin{equation}
t_{\texttt{Trig}} = \arctan\left(\frac{t_{\texttt{FM}}}{1-t_{\texttt{FM}}}\right), \qquad
\bx_{t,\texttt {Trig }} = \frac{\bx_{t,\texttt{FM}}}{\sqrt{t_{\texttt{FM}}^2 + (1-t_{\texttt{FM}})^2}}.
\end{equation}
The TrigFlow model $\boldsymbol{F_{\theta}}$ processes this transformed input, and its output is then converted back into the velocity field estimate $\boldsymbol{\widehat{v}_{\theta}}(\bx_{t,\texttt{FM}}, t_{\texttt{FM}}, \boldsymbol{y})$ required by the Flow Matching framework. This output transformation is defined by:
\begin{equation}
\boldsymbol{\widehat{v}_{\theta}}(\bx_{t,\texttt{FM}}, t_{\texttt{FM}}, \boldsymbol{y}) = \frac{1}{\sqrt{t_{\texttt{FM}}^2 + (1-t_{\texttt{FM}})^2}}
\boldsymbol{F_{\theta}}\!\left(\bx_{t,\texttt{Trig}}, t_{\texttt{Trig}}, \boldsymbol{y}\right)
- \frac{1-2t_{\texttt{FM}}}{t_{\texttt{FM}}^2 + (1-t_{\texttt{FM}})^2} \bx_{t, \texttt{FM}}.
\end{equation}
This bidirectional conversion ensures complete interoperability, allowing practitioners to flexibly combine models and samplers from either framework.

\subsection{sCM v.s. MeanFlow: MeanFlow as a Synchronization Limit of sCM}


Under the Flow Matching (FM) parameterization, where the velocity field is directly modeled by a neural network, i.e., $\boldsymbol{v}_\theta(\bx_t, t) = \boldsymbol{F}_\theta(\bx_t, t)$, a direct comparison between the loss gradients of Stochastic Control Matching (sCM) and our proposed MeanFlow (MF) reveals an insightful connection. The respective gradients with respect to the network parameters $\theta$, denoted as $\nabla_\theta \mathcal{L}'$, are given by:
\begin{align}
    \nabla_\theta \mathcal{L}_{\text{sCM}}' &= \mathbb{E}_{t, \bx_t} \left[ -\left\langle \nabla_\theta \boldsymbol{F}_\theta(\bx_t, t), \boldsymbol{v}_t - \boldsymbol{F}_{\theta^{-}}(\bx_t, t) - t \frac{d\boldsymbol{F}_{\theta^{-}}}{dt}(\bx_t, t) \right\rangle \right] \label{eq:scm_grad} \\
    \nabla_\theta \mathcal{L}_{\text{MF}}' &= \mathbb{E}_{t, \bx_t} \left[ -\left\langle \nabla_\theta \boldsymbol{F}_\theta(\bx_t, t), \boldsymbol{v}_t - \boldsymbol{F}_{\theta}(\bx_t, t) - t \frac{d\boldsymbol{F}_{\theta}}{dt}(\bx_t, t) \right\rangle \right] \label{eq:mf_grad}
\end{align}
A comparison of Eq.~\eqref{eq:scm_grad} and Eq.~\eqref{eq:mf_grad} highlights their nearly identical structure. The sole distinction lies in the regression target provided for the network's output within the inner product. Specifically, sCM employs a target network $\boldsymbol{F}_{\theta^{-}}$, parameterized by $\theta^{-}$, which is typically an exponential moving average (EMA) of the online network's parameters $\theta$. This technique is widely used to stabilize training by providing a more consistent and slowly-evolving target.

In stark contrast, MeanFlow utilizes the current online network $\boldsymbol{F}_{\theta}$ for its entire regression target. Consequently, the MeanFlow gradient is mathematically equivalent to the sCM gradient under the condition that the target network is fully synchronized with the online network at every training step, i.e., by setting $\theta^{-} = \theta$. This interpretation frames MeanFlow not as a distinct method, but as a specific, simplified variant of sCM that dispenses with the EMA-based target stabilization. This design choice results in a more up-to-date, self-referential training dynamic, differentiating its behavior from standard sCM.

\subsection{CM v.s. IMM: Consistency Models as a Special Case of Implicit Moment Matching}

As shown in the Eq.(12) of the IMM paper (\cite{zhou2025inductive}), the IMM loss is:

\begin{align}\label{eq:imm_simple_obj}
  & \mathcal{L}_{\text{IMM}} (\theta) = \mathbb{E}_{\boldsymbol{x}_t, \boldsymbol{x}_t', \boldsymbol{x}_r, \boldsymbol{x}_r',s, t}  \Big[w(s,t)
  \Big[ k \Big(\boldsymbol{y}_{s, t}, \boldsymbol{y}_{s, t}'\Big)
  + k\Big({\boldsymbol{y}}_{s, r},{\boldsymbol{y}}_{s, r}'\Big)  - k\Big(\boldsymbol{y}_{s, t}, {\boldsymbol{y}}_{s, r}'\Big) - k\Big(\boldsymbol{y}_{s, t}', {\boldsymbol{y}}_{s, r}\Big)  \Big] \Big],
\end{align}
where
$\boldsymbol{y}_{s, t} = \boldsymbol{f}_{s, t}^{\theta}(\boldsymbol{x}_t)$,
$\boldsymbol{y}_{s, t}' = \boldsymbol{f}_{s, t}^{\theta}(\boldsymbol{x}_t')$, 
${\boldsymbol{y}}_{s, r} = \boldsymbol{f}_{s, r}^{\theta^-}(\boldsymbol{x}_r)$,
${\boldsymbol{y}}_{s, r}' = \boldsymbol{f}_{s, r}^{\theta^-}(\boldsymbol{x}_r')$, $k(\cdot, \cdot)$ is a kernel function, and $w(s,t)$ is a prior weighting function.
As is illustrated in Appendix G of the original paper, when we set
$\boldsymbol{x}_t=\boldsymbol{x}_t'$, $\boldsymbol{x}_r=\boldsymbol{x}_r'$, we have $\boldsymbol{f}_{s,t}^\theta(\boldsymbol{x}_t) = \boldsymbol{f}_{s,t}^\theta(\boldsymbol{x}_t')$
and $\boldsymbol{f}_{s,r}^{\theta^-}(\boldsymbol{x}_r) = \boldsymbol{f}_{s,r}^{\theta^-}(\boldsymbol{x}_r')$ and let $k(x,y)=-||x-y||^2$,
which means $\boldsymbol{y}_{s, t} = \boldsymbol{y}_{s, t}'$ and ${\boldsymbol{y}}_{s, r} = {\boldsymbol{y}}_{s, r}'$.
So $k(\boldsymbol{y}_{s, t}, \boldsymbol{y}_{s, t}')=k(\boldsymbol{f}_{s,t}^\theta(\boldsymbol{x}_t), \boldsymbol{f}_{s,t}^\theta(\boldsymbol{x}_t'))=0$ and $ k(\boldsymbol{y}_{s, r}, \boldsymbol{y}_{s, r}') = k(\boldsymbol{f}_{s,r}^\theta(\boldsymbol{x}_r), \boldsymbol{f}_{s,r}^\theta(\boldsymbol{x}_r')) =0$ by definition. The original IMM loss in \cref{eq:imm_simple_obj} reduce to:
\begin{align}\label{eq:imm_simple_obj2}
  & \mathcal{L}_{\text{IMM}} (\theta) = \mathbb{E}_{\boldsymbol{x}_t, \boldsymbol{x}_r, s, t}  \Big[w(s,t)
  \Big[ 
   - k\Big(\boldsymbol{y}_{s, t}, {\boldsymbol{y}}_{s, r}'\Big)
   - k\Big(\boldsymbol{y}_{s, t}', {\boldsymbol{y}}_{s, r}\Big)  \Big] \Big].
\end{align}

Furthermore, since we use $k(x,y)=-||x-y||^2$, 
$- k(\boldsymbol{y}_{s, t}, {\boldsymbol{y}}_{s, r}') = ||\boldsymbol{f}_{s, t}^{\theta}(\boldsymbol{x}_t) - \boldsymbol{f}_{s, r}^{\theta^-}(\boldsymbol{x}_r')||^2 = ||\boldsymbol{f}_{s, t}^{\theta}(\boldsymbol{x}_t) - \boldsymbol{f}_{s, r}^{\theta^-}(\boldsymbol{x}_r)||^2$, 
and $- k(\boldsymbol{y}_{s, t}', {\boldsymbol{y}}_{s, r}) = ||\boldsymbol{f}_{s, t}^{\theta}(\boldsymbol{x}_t') - \boldsymbol{f}_{s, r}^{\theta^-}(\boldsymbol{x}_r)||^2 = ||\boldsymbol{f}_{s, t}^{\theta}(\boldsymbol{x}_t) - \boldsymbol{f}_{s, r}^{\theta^-}(\boldsymbol{x}_r)||^2$, a more simplified expression is
\begin{align}\label{eq:imm_simple_obj3}
  & \mathcal{L}_{\text{IMM}} (\theta) = \mathbb{E}_{\boldsymbol{x}_t, \boldsymbol{x}_r, s, t}  \Big[2w(s,t)
  \Big[ 
   ||\boldsymbol{f}_{s, t}^{\theta}(\boldsymbol{x}_t) - \boldsymbol{f}_{s, r}^{\theta^-}(\boldsymbol{x}_r)||^2  \Big] \Big].
\end{align}

If $s$ is a small positive constant, we further have $\boldsymbol{f}_{s,t}^\theta(\boldsymbol{x}_t)\approx \boldsymbol{g}_\theta(\boldsymbol{x}_t,t)$ , where $\boldsymbol{g}_\theta(\boldsymbol{x}_t,s,t)$ is the diffusion model parameterized with EDM (\cite{karras2022elucidating}) and we drop $s$ as input. If $\boldsymbol{g}_\theta(\boldsymbol{x}_t,t)$ itself satisfies boundary condition at $s=0$, we can directly take $s=0$ in which case $\boldsymbol{f}_{0,t}^\theta(\boldsymbol{x}_t) = \boldsymbol{g}_\theta(\boldsymbol{x}_t,t)$. And under these assumptions, and omit the constant 2, our loss becomes
\begin{align}\label{eq:imm_simple_obj4}
  & \mathcal{L}_{\text{IMM}} (\theta) = \mathbb{E}_{\boldsymbol{x}_t, \boldsymbol{x}_r, t}  \Big[w(t)
  \Big[ 
   ||\boldsymbol{g}_{\theta}(\boldsymbol{x}_t, t) - \boldsymbol{g}_{\theta^-}(\boldsymbol{x}_r,r)||^2  \Big] \Big],
\end{align}
which is a discrete-time consistency model(CM) loss using $\ell_2$ distance. In short, the IMM loss degenerates to the discrete-time Consistency Model (CM) loss when utilizing a single-particle estimate (setting $\boldsymbol{x}_t=\boldsymbol{x}_t'$ and $\boldsymbol{x}_r=\boldsymbol{x}_r'$), employing the negative squared Euclidean distance as the kernel function (i.e., $k(x,y)=-\|x-y\|^2$), and setting the target time $s$ to 0.

%% file: sec/e4.tex
\section{Text-to-Image Adaptation with \textsc{FLUX.1-lite}}

In this section, we detail the adaptation of the 8B \textsc{FLUX.1-lite} (\cite{fluxlite}) model for text-to-image generation. We introduce and evaluate two primary methodologies: an adaptation based on a \textbf{simplified Consistency Model (sCM)} objective and a novel \textbf{MeanFlow} training objective.

\subsection{Experimental Setup}

For all subsequent experiments, we utilize 32 Nvidia H20 GPUs for training. The dataset employed is a proprietary high-quality text-to-image dataset. While the dataset cannot be publicly released, it ensures a consistent and fair comparison across all experiments. We adopt the \textsc{FLUX.1-lite} model with 8 billion parameters as the teacher and aim to distill a few-step student model from it. For sCM, we distill the student model for 3,000 iterations, as we observed no further improvement in the GenEval overall score beyond this point. Conversely, for MeanFlow, we report results at 25,000 iterations, as extending the training duration yielded continuous gains in the GenEval score.

\subsection{Methodological Adaptation for Text-to-Image Generation}

\subsubsection{sCM Adaptation}\textbf{Normalizing DiT timestep inputs stabilizes training.} During the training of sCM using \textsc{FLUX.1-lite}, we observed a phenomenon similar to that reported in prior works (\cite{lu2025simp,chen2025sana}). Specifically, when the Diffusion Transformer's timestep input ranges from 0 to 1000, the gradient norm increases continuously during training, eventually leading to collapse. To mitigate this, we first rescaled the diffusion transformer's timestep to the range $[0, 1]$. We treated the original \textsc{FLUX.1-lite} as the teacher and initialized a student model with an identical structure, configured to accept timesteps in $[0, 1]$. We distilled the teacher into the student using identical images, text prompts, and proportionally scaled timesteps by minimizing the smooth L1 loss between their outputs. This distillation was performed at a resolution of $1024 \times 1024$ with a total batch size of $128$ for $120$K iterations. The resulting student model achieved performance comparable to the teacher on the GenEval benchmark, as shown in the first two lines of \cref{tab:geneval}.

\textbf{Guiding the student model with the teacher's velocity prediction.} 
Subsequently, we utilized this timestep-rescaled model ($t \in [0, 1]$) to train the sCM. Instead of the standard consistency training paradigm, we employed consistency distillation, which uses the classifier-free guided output of the teacher as the velocity target. The training was conducted at a resolution of $512$ with a total batch size of $128$ and a learning rate of $1\text{e-}6$.

\subsubsection{MeanFlow Adaptation}

\textbf{Dual-timestep input mechanism.} 
As MeanFlow models the average velocity between timesteps $r$ and $t$, the diffusion transformer requires an additional time input. The standard \textsc{FLUX.1-lite} encodes a single timestep $t$ using sinusoidal embeddings and an MLP projection for AdaLN modulation. We replicate this embedding branch—initialized with the same weights—to process the time differential $(t-r)$. The outputs of the original branch (encoding $t$) and the cloned branch (encoding $t-r$) are summed before being passed to the AdaLN layers.

\textbf{Teacher Distillation for MeanFlow.} 
While MeanFlow can be formulated as a standalone training objective (Algorithm~\ref{alg:meanflowc2i}), we empirically find it significantly more effective as a distillation technique (Algorithm~\ref{alg:meanflowt2i}). The primary distinction lies in the target velocity. In the standard training formulation, the target $v = e - x$ serves as an unbiased but highly stochastic estimator of the true vector field, introducing substantial variance into the optimization of $u$. In contrast, Algorithm~\ref{alg:meanflowt2i} employs a pre-trained flow-matching teacher $g_n$ to provide the target $v_{\text{teacher}} = g_n(z, t)$. We utilize the teacher's instantaneous velocity as a direct guidance signal because the teacher has already converged to the conditional expectation of the vector field, i.e., $g_n(z, t) \approx \mathbb{E}[v | z]$. Consequently, $v_{\text{teacher}}$ represents a denoised, deterministic approximation of the optimal transport path. By substituting the noisy raw target with this stable teacher signal, the MeanFlow objective $u_{\text{tgt}} = v_{\text{teacher}} - (t - r) \frac{du}{dt}$ focuses purely on rectifying the curvature of the trajectory via the Jacobian correction term, rather than learning the data distribution from scratch.

\textbf{High-Order Loss Achieves Better Performance.}
While the original MeanFlow paper \cite{geng2025mean} suggests that $p=1$ or $p=0.5$ are generally optimal in the loss function $\mathcal{L} = \|\Delta\|_2^{2\gamma}$ (where $\Delta$ denotes the regression error), we empirically find that setting $\gamma=2$ yields superior performance. This configuration, which effectively minimizes the fourth power of the norm, imposes a distinct gradient behavior beneficial for our distillation objective. Quantitatively, this adjustment leads to a substantial boost on the vanilla MeanFlow target: the GenEval score increases from 44.04\% to 48.65\%.

\textbf{Improved CFG for MeanFlow in Text-to-Image.}
We compare the vanilla MeanFlow (where $\kappa=0$) with the Improved CFG variant introduced in Appendix B.2 of the MeanFlow paper (\cite{geng2025mean}). The improved version incorporates a mixing scale $\kappa$ to blend class-conditional and unconditional predictions in the regression target, aligning more closely with standard practices in classifier-free guidance. Experiments show that this modification is beneficial for text-to-image generation, boosting the GenEval score from 48.65\% to 51.41\%.

\subsection{Quantitative and Qualitative Results}

\begin{table}[h!]
\centering
\caption{\textbf{Quantitative comparison on the GenEval benchmark.} We evaluate the original \textsc{FLUX.1-lite}, the timestep-rescaled teacher, and the distilled students (sCM and MeanFlow) across varying Numbers of Function Evaluations (NFE). sCM maintains robust performance at NFE=1/2, while MeanFlow requires NFE=4 to match the teacher's quality.}
\label{tab:geneval}
\begin{subtable}{0.99\textwidth}
    \centering
    \begin{tabular}{lcccccccc}
    \toprule
     & NFE & Single Obj. & Two Obj. & Counting & Colors & Position & Color Attri. & Overall \\
    \midrule
    \textsc{FLUX.1-lite} & 28 & 98.44 & 65.40 & 29.69 & 78.19 & 13.50 & 36.25 & 53.58 \\
    Rescaled & 28 & 97.81 & 63.89 & 39.06 & 74.73 & 12.50 & 33.50 & 53.58 \\
    \midrule
    sCM & 4 & 97.19 & 56.82 & 35.62 & 76.86 & 8.75 & 38.50 & 52.29 \\
    sCM & 2 & 94.69 & 55.81 & 38.44 & 74.20 & 10.25 & 43.50 & 52.81 \\
    sCM & 1 & 89.69 & 36.87 & 26.56 & 67.55 & 6.50 & 32.50 & 43.28 \\
    \midrule
    MeanFlow & 4 & 96.25 & 64.14 & 38.44 & 69.15 & 13.00 & 27.50 & 51.41 \\
    MeanFlow & 2 & 61.88 & 31.31 & 18.12 & 39.89 & 7.25 & 18.50 & 29.49 \\
    MeanFlow & 1 & 2.81 & 0.25 & 0.31 & 0.80 & 0.25 & 0.25 & 0.78 \\
    \bottomrule
    \end{tabular}
\end{subtable}
\end{table}

\begin{table}[h!]
\centering
\caption{\textbf{Quantitative results on DPG-Bench.} This benchmark evaluates the alignment of global structure, entities, attributes, and relations. MeanFlow achieves teacher-level performance at NFE=4, whereas sCM demonstrates superior stability and alignment at lower step counts (NFE $\le$ 2).}
\label{tab:dpgbench}
\begin{subtable}{0.99\textwidth}
    \centering
    \begin{tabular}{lccccccc}
    \toprule
    & NFE & Global & Entity & Attribute & Relation & Other & Overall \\
    \midrule
    \textsc{FLUX.1-lite} & 28 & 86.09 & 88.73 & 86.27 & 88.50 & 80.87 & 80.20 \\
    Rescaled & 28 & 83.93 & 87.00 & 86.0 & 88.86 & 86.54 & 79.83 \\
    \midrule
    sCM & 4 & 82.21 & 86.28 & 85.49 & 88.41 & 86.03 & 77.85 \\
    sCM & 2 & 82.22 & 87.26 & 85.90 & 88.66 & 86.50 & 79.06 \\
    sCM & 1 & 82.33 & 82.67 & 85.48 & 83.66 & 81.74 & 75.12 \\
    \midrule
    MeanFlow & 4 & 90.83 & 86.15 & 87.20 & 88.27 & 86.35 & 80.03 \\
    MeanFlow & 2 & 79.81 & 80.50 & 83.67 & 84.13 & 80.49 & 71.09 \\
    MeanFlow & 1 & 52.44 & 48.90 & 59.83 & 50.10 & 54.39 & 27.55 \\
    \bottomrule
    \end{tabular}
\end{subtable}
\end{table}

\textbf{Quantitative Comparison.} 
We report the evaluation results on GenEval (Table~\ref{tab:geneval}) and DPG-Bench (Table~\ref{tab:dpgbench}). 
First, the \textbf{Rescaled} teacher model exhibits performance nearly identical to the original \textsc{FLUX.1-lite}, validating that normalizing the timestep range to $[0, 1]$ does not compromise generation quality.

Comparing the distillation methods reveals a distinct trade-off between step efficiency and peak performance. First, sCM excels in extreme few-step regimes. sCM demonstrates remarkable stability at NFE=2 and even NFE=1. On GenEval, sCM (NFE=2) achieves an overall score of 52.81\%, effectively matching the teacher's 53.58\%. Even at a single step (NFE=1), it retains a respectable score of 43.28\%. A similar trend is observed on DPG-Bench, where sCM maintains high alignment scores across all step counts. We also find that MeanFlow requires sufficient steps but achieves high fidelity. At NFE=4, it outperforms sCM on DPG-Bench (80.03 vs. 77.85) and nearly matches the teacher, suggesting that its trajectory straightening is highly effective when given sufficient integration steps. However, performance collapses at NFE=1 and NFE=2 (\eg,  0.78\% on GenEval at NFE=1), indicating that the learned vector field, while straight, still requires a minimum number of discretization steps to traverse accurately.

\begin{figure}[t]
    \centering
    \includegraphics[width=0.99\linewidth]{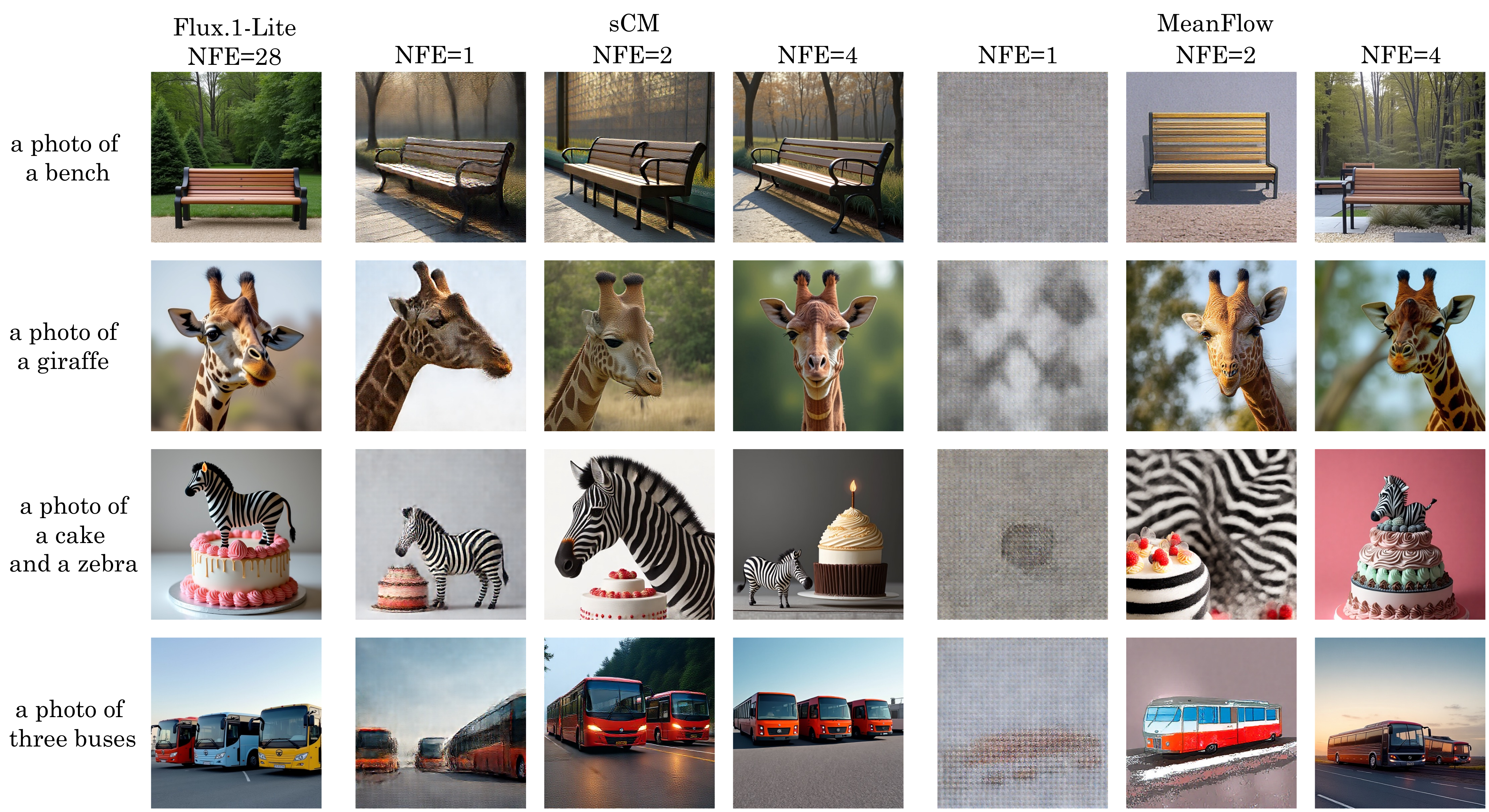} 
    \caption{\textbf{Qualitative comparison of distillation methods.} We visualize samples generated by the teacher (left) and the two student models at NFE=\{1, 2, 4\}. \textbf{sCM} produces structurally coherent images even at a single step. In contrast, \textbf{MeanFlow} exhibits generation collapse at NFE=1 (pure noise) and requires NFE=4 to converge to high-fidelity results.}
    \label{fig:vis}
\end{figure}

\textbf{Qualitative Visualization.}
These quantitative findings are visually corroborated in Figure~\ref{fig:vis}. For sCM, as shown in the \texttt{sCM }column, the model produces structurally coherent images even at NFE=1 (\eg,  the giraffe and the buses are clearly recognizable). Increasing the steps to 2 or 4 primarily refines high-frequency details and textures. The \texttt{MeanFlow} column illustrates the collapse at NFE=1, resulting in pure noise (gray outputs). At NFE=2, the model begins to form semantic content (\eg,  the bench appears), but significant artifacts and noise remain (visible in the giraffe and zebra examples). However, at NFE=4, MeanFlow produces images with exceptional sharpness and correct semantics, often surpassing sCM in fine-grained detail (\eg,  the fur texture of the giraffe and the reflection on the buses). In summary, sCM is the optimal choice for real-time applications requiring NFE $\le$ 2, while MeanFlow is preferable for scenarios where a slightly higher budget (NFE=4) is acceptable for maximum quality.

%% file: sec/e5.tex
\section{Codebase}

Our codebase is built upon the Hugging Face Diffusers library and implements multiple distillation algorithms for accelerating the \textsc{FLUX.1-lite} text-to-image diffusion model. The implementation includes two primary training pipelines: (1) \textbf{MeanFlow}, a trajectory distillation approach utilizing Jacobian-vector products (JVPs) to enforce consistency along the flow path with optional classifier-free guidance integration; and (2) \textbf{simplified Consistency Matching (sCM)}, which leverages tangent vector matching with learned per-sample variance through a TrigFlow reparameterization of the \textsc{FLUX.1-lite} model.

The framework employs a teacher-student paradigm based on a modified \textsc{FLUX.1-lite} MMDiT architecture. We utilize DeepSpeed ZeRO with bfloat16 mixed-precision and gradient checkpointing for efficient distributed training. The pipeline performs on-the-fly encoding via the pretrained AutoencoderKL, operating in a 16-channel latent space with 8$\times$ downsampling.

%% file: sec/app.tex
\appendix

\section{Proofs}

\subsection{Interconversion between TrigFlow and Flow Matching}

The following derivation is based on the supplementary materials of sana-sprint~\cite{chen2025sana}, with some modifications made for easier understanding.
The TrigFlow framework defines the noisy input sample as:
\begin{equation}
\bx_{t,\texttt{Trig}} = \cos(t_{\texttt{Trig}}) \bx_0 + \sin(t_{\texttt{Trig}}) \boldsymbol{z}.
\end{equation}

The Signal-to-Noise Ratios (SNRs) for the Flow Matching and TrigFlow models are defined respectively as:
\begin{equation}
\text{SNR}(t_{\texttt{FM}}) = (\frac{1-t_{\texttt{FM}}}{t_{\texttt{FM}}})^2, \quad \text{SNR}(t_{\texttt{Trig}}) = (\frac{\cos(t_{\texttt{Trig}})}{\sin(t_{\texttt{Trig}})})^2 = (\frac{1}{\tan(t_{\texttt{Trig}})})^2.
\end{equation}
To align the models, we match their SNRs. We seek the corresponding time $t_{\texttt{FM}}$ in the Flow Matching framework that satisfies:
\begin{equation}
(\frac{1-t_{\texttt{FM}}}{t_{\texttt{FM}}})^2 = (\frac{1}{\tan(t_{\texttt{Trig}})})^2.
\end{equation}
Solving this equation yields the interconversion relationship between $t_{\texttt{FM}}$ and $t_{\texttt{Trig}}$:
\begin{equation}
t_{\texttt{FM}} = \frac{\sin{(t_{\texttt{Trig}})}}{\sin{(t_{\texttt{Trig}})} + \cos{(t_{\texttt{Trig}})}}, \quad t_{\texttt{Trig}} = \arctan{(\frac{t_{\texttt{FM}}}{1-t_{\texttt{FM}}})}.
\end{equation}

From this relationship, we can also derive the following three useful identities:
\begin{equation}
\cos(t_{\texttt{Trig}}) = \cos(\arctan{(\frac{t_{\texttt{FM}}}{1-t_{\texttt{FM}}})}) = \frac{1-t_{\texttt{FM}}}{ \sqrt{t_{\texttt{FM}}^2 + (1-t_{\texttt{FM}})^2}},
\end{equation}

\begin{equation}
\sin(t_{\texttt{Trig}}) = \sin(\arctan{(\frac{t_{\texttt{FM}}}{1-t_{\texttt{FM}}})}) = \frac{t_{\texttt{FM}}}{ \sqrt{t_{\texttt{FM}}^2 + (1-t_{\texttt{FM}})^2}},
\end{equation}

\begin{equation}
\cos(t_{\texttt{Trig}}) + \sin(t_{\texttt{Trig}}) = \frac{1}{ \sqrt{t_{\texttt{FM}}^2 + (1-t_{\texttt{FM}})^2}}.
\end{equation}

The conversion relationship between the noisy samples $\bx_{t, \texttt{FM}}$ and $\bx_{t, \texttt{Trig}}$ can then be expressed as:
\begin{equation}
\begin{aligned}
\bx_{t, \texttt{FM}} &= (1-t_{\texttt{FM}})\bx_0 + t_{\texttt{FM}} \boldsymbol{z} \\
&= \sqrt{t_{\texttt{FM}}^2 + (1-t_{\texttt{FM}})^2} \cdot (\frac{1 - t_{\texttt{FM}}}{ \sqrt{t_{\texttt{FM}}^2 + (1-t_{\texttt{FM}})^2}} \cdot \bx_0 + \frac{t_{\texttt{FM}}}{ \sqrt{t_{\texttt{FM}}^2 + (1-t_{\texttt{FM}})^2}} \cdot \boldsymbol{z}), \\
&= \sqrt{t_{\texttt{FM}}^2 + (1-t_{\texttt{FM}})^2} \cdot [\cos(t_{\texttt{Trig}})\bx_0 + \sin(t_{\texttt{Trig}}) \boldsymbol{z}], \\
&= \sqrt{t_{\texttt{FM}}^2 + (1-t_{\texttt{FM}})^2} \cdot \bx_{t, \texttt{Trig}}, \\
&= \frac{1}{ \cos(t_{\texttt{Trig}}) + \sin(t_{\texttt{Trig}}) } \cdot \bx_{t, \texttt{Trig}},
\end{aligned}
\end{equation}

This conversion confirms that the samples from both frameworks can be mapped to the same distribution, fulfilling our objective. Our next goal is to determine the optimal estimator for the TrigFlow model, $\boldsymbol{F_{\theta}}$, based on the optimal estimator of the Flow Matching model, $\boldsymbol{v_{\theta}}(\bx_{t, \texttt{FM}}, t_{\texttt{FM}}, \boldsymbol{y})$. We begin by considering an ideal scenario assuming sufficient model capacity. In this optimal setting, the flow matching model's solution is:
\begin{equation}
\boldsymbol{v}^*(\bx_{t, \texttt{FM}}, t_{\texttt{FM}}, \boldsymbol{y}) = \mathbb{E}[\boldsymbol{z}-\bx_0 | \bx_{t_{\texttt{FM}}}, \boldsymbol{y}],
\end{equation}
This is the optimal solution as conditional expectation minimizes the Mean Squared Error (MSE) loss. Analogously, the optimal solution for the TrigFlow model is given by:
\begin{equation}
\boldsymbol{F}^*(\bx_{t, \texttt{Trig}}, t_{\texttt{Trig}}, \boldsymbol{y}) = \mathbb{E}[\cos{(t_{\texttt{Trig}})}\boldsymbol{z}-\sin{(t_{\texttt{Trig}})}\bx_0 | \bx_{t_{\texttt{Trig}}}, \boldsymbol{y}].
\end{equation}

We now leverage the linearity of conditional expectation to derive the relationship:
\begin{equation}
\begin{aligned}
& \mathbb{E}[ \cos{(t_{\texttt{Trig}})}\boldsymbol{z} - \sin{(t_{\texttt{Trig}})}\bx_0 | x_{t_{\texttt{Trig}}}, \boldsymbol{y}].\\
=& \mathbb{E}[ \frac{1-t_{\texttt{FM}}}{\sqrt{t_{\texttt{FM}}^2 + (1-t_{\texttt{FM}})^2}} \boldsymbol{z} - \frac{t_{\texttt{FM}}}{\sqrt{t_{\texttt{FM}}^2 + (1-t_{\texttt{FM}})^2}}\bx_0| \bx_{t_{\texttt{FM}}}, \boldsymbol{y}] \\
=& \frac{1-2t_{\texttt{FM}}}{\sqrt{t_{\texttt{FM}}^2 + (1-t_{\texttt{FM}})^2}} \mathbb{E}[(1-t_{\texttt{FM}})\cdot\bx_0 + t_{\texttt{FM}}\cdot\boldsymbol{z}| \bx_{t_{\texttt{FM}}}, \boldsymbol{y}]
+ \frac{t_{\texttt{FM}}^2 + (1-t_{\texttt{FM}})^2}{\sqrt{t_{\texttt{FM}}^2 + (1-t_{\texttt{FM}})^2}}\mathbb{E}[\boldsymbol{z}-\bx_0 | \bx_{t_{\texttt{FM}}}, \boldsymbol{y}]\\
=&\frac{1-2t_{\texttt{FM}}}{\sqrt{t_{\texttt{FM}}^2 + (1-t_{\texttt{FM}})^2}}\cdot \bx_{t_{\texttt{FM}}}
+ \frac{t_{\texttt{FM}}^2 + (1-t_{\texttt{FM}})^2}{\sqrt{t_{\texttt{FM}}^2 + (1-t_{\texttt{FM}})^2}}\mathbb{E}[\boldsymbol{z}-\bx_0 | \bx_{t_{\texttt{FM}}}, \boldsymbol{y}]\\
= &[\cos(t_{\texttt{Trig}}) - \sin(t_{\texttt{Trig}})] \cdot \bx_{t_{\texttt{FM}}}
+\frac{1}{\cos(t_{\texttt{Trig}}) + \sin(t_{\texttt{Trig}})} \cdot \mathbb{E}[\boldsymbol{z}-\bx_0 | \bx_{t_{\texttt{FM}}}, \boldsymbol{y}]\\
= &\frac{\cos(t_{\texttt{Trig}}) - \sin(t_{\texttt{Trig}})}{\cos(t_{\texttt{Trig}}) + \sin(t_{\texttt{Trig}})} \cdot \bx_{t, \texttt{Trig}}
+\frac{1}{\cos(t_{\texttt{Trig}}) + \sin(t_{\texttt{Trig}})} \cdot \mathbb{E}[\boldsymbol{z}-\bx_0 | \bx_{t_{\texttt{FM}}}, \boldsymbol{y}]\\
\end{aligned}
\end{equation}

This derivation ultimately yields the following conversion formulas:
\begin{equation}
\boldsymbol{F}^*(\bx_{t, \texttt{Trig}}, t_{\texttt{Trig}}, \boldsymbol{y})
= \frac{\cos(t_{\texttt{Trig}}) - \sin(t_{\texttt{Trig}})}{\cos(t_{\texttt{Trig}}) + \sin(t_{\texttt{Trig}})} \cdot \bx_{t, \texttt{Trig}} + \frac{1}{\cos(t_{\texttt{Trig}}) + \sin(t_{\texttt{Trig}})} \cdot \boldsymbol{v}^*(\bx_{t, \texttt{FM}}, t_{\texttt{FM}}, \boldsymbol{y}), \\
\end{equation}

\begin{equation}
\boldsymbol{v}^*(\bx_{t, \texttt{FM}}, t_{\texttt{FM}}, \boldsymbol{y})
= 
\frac{2t_{\texttt{FM}}-1}{t_{\texttt{FM}}^2 + (1-t_{\texttt{FM}})^2}
\cdot \bx_{t, \texttt{FM}} + \frac{1}{\sqrt{t_{\texttt{FM}}^2 + (1-t_{\texttt{FM}})^2}} \cdot \boldsymbol{F}^*(\bx_{t, \texttt{Trig}}, t_{\texttt{Trig}}, \boldsymbol{y}), \\
\end{equation}

\subsection{sCM v.s. MeanFlow: Equivalence of Gradients}

This derivation demonstrates that under the Flow Matching parameterization, the training objective gradients for the simplified consistency model (sCM) and a simplified version of Meanflow are structurally equivalent. Instead of the TrigFlow, we use the simpler Flow Matching (FM) parameterization for the time-dependent function $\boldsymbol{f}_\theta(\boldsymbol{x}_t, t) = \boldsymbol{x}_t - t \boldsymbol{F}_\theta(\boldsymbol{x}_t, t)$, which satisfies the boundary condition $\boldsymbol{f}_\theta(\boldsymbol{x}_0, 0) = \boldsymbol{x}_0$. The time derivative of $\boldsymbol{f}_\theta(\boldsymbol{x}_t, t)$ is:
\begin{equation}
\frac{\mathrm{d}}{\mathrm{d}t}\boldsymbol{f}_\theta(\boldsymbol{x}_t, t) = \boldsymbol{v}_t - \boldsymbol{F}_\theta(\boldsymbol{x}_t, t) - t \frac{\mathrm{d}\boldsymbol{F}_\theta(\boldsymbol{x}_t, t)}{\mathrm{d}t},
\end{equation}
where $\boldsymbol{v}_t = \frac{\mathrm{d}\boldsymbol{x}_t}{\mathrm{d}t}$ is the velocity of the path $\boldsymbol{x}_t$. Under this parameterization, the loss of a continuous-time consistency model is (as derived in sCM~\cite{lu2025simp}): 

\begin{equation}
\mathcal{L}_{\text{sCM}} = \mathbb{E}_{t, \boldsymbol{x}_t} \left[ \boldsymbol{f}_\theta^{\top}\left(\boldsymbol{x}_t, t\right) \frac{\mathrm{d} \boldsymbol{f}_{\theta^{-}}\left(\boldsymbol{x}_t, t\right)}{\mathrm{d} t} \right],
\end{equation}
where $\theta^{-}$ denotes the parameters with stopped gradients. This can be written as:
\begin{equation}
\mathcal{L}_{\text{sCM}} = \mathbb{E}_{t, \boldsymbol{x}_t} \left[ \left\langle \boldsymbol{f}_\theta(\boldsymbol{x}_t, t), \boldsymbol{v}_t - \boldsymbol{F}_{\theta^{-}}(\boldsymbol{x}_t, t) - t \frac{\mathrm{d}\boldsymbol{F}_{\theta^{-}}}{\mathrm{d}t} \right\rangle \right].
\end{equation}

To find the gradient for optimization, we differentiate this loss with respect to $\theta$. Noting that $\nabla_\theta \boldsymbol{f}_\theta(\boldsymbol{x}_t, t) = -t \nabla_\theta \boldsymbol{F}_\theta(\boldsymbol{x}_t, t)$, this yields:
\begin{equation}
\nabla_\theta \mathcal{L}_{\text{sCM}} = \mathbb{E}_{t, \boldsymbol{x}_t} \left[ - \left\langle t \nabla_\theta \boldsymbol{F}_\theta(\boldsymbol{x}_t, t), \boldsymbol{v}_t - \boldsymbol{F}_{\theta^{-}}(\boldsymbol{x}_t, t) - t \frac{\mathrm{d}\boldsymbol{F}_{\theta^{-}}}{\mathrm{d}t} \right\rangle \right].
\end{equation}

On the other hand, the MeanFlow loss is (as derived in MeanFlow~\cite{geng2025mean}):
\begin{equation}
\begin{aligned}
&\mathcal{L}_{\text{MeanFlow}}(\theta) =\mathbb{E}_{t, \boldsymbol{z}_t} \left[ \left\| {\boldsymbol{u}}_\theta\left(\boldsymbol{z}_t, r, t\right)-\operatorname{sg}\left(\boldsymbol{u}_{\text {tgt }}\right)\right\|_2^2 \right], \\
&\text { where } \! \boldsymbol{u}_{\text {tgt }} = \boldsymbol{v}_t-(t-r)\left(\boldsymbol{v}_t \partial_{\boldsymbol{z}} {\boldsymbol{u}}_\theta+\partial_t {\boldsymbol{u}}_\theta\right).
\end{aligned}
\end{equation}

When we set $r$ to 0, it becomes:
\begin{equation}
\begin{aligned}
&\mathcal{L}_{\text{MeanFlow}}(\theta) =\mathbb{E}_{t, \boldsymbol{z}_t} \left[ \left\| {\boldsymbol{u}}_\theta\left(\boldsymbol{z}_t, t\right)-\operatorname{sg}\left(\boldsymbol{u}_{\text {tgt }}\right)\right\|_2^2 \right], \\
&\text { where } \! \boldsymbol{u}_{\text {tgt }} = \boldsymbol{v}_t-t\left(\boldsymbol{v}_t \partial_{\boldsymbol{z}} {\boldsymbol{u}}_\theta+\partial_t {\boldsymbol{u}}_\theta\right).
\end{aligned}
\end{equation}

After unifying the notation, $\boldsymbol{z}_t \Rightarrow \boldsymbol{x}_t$, $\boldsymbol{u}_\theta \Rightarrow \boldsymbol{F}_\theta$, and identifying the total derivative term $\boldsymbol{v}_t \partial_{\boldsymbol{z}} \boldsymbol{u}_\theta+\partial_t {\boldsymbol{u}}_\theta \Rightarrow \frac{\mathrm{d} \boldsymbol{F}_{\theta^{-}}(\boldsymbol{x}_t, t)}{\mathrm{d}t}$ (where the stop-gradient operator $\operatorname{sg}(\cdot)$ implies the use of $\theta^{-}$), the MeanFlow loss becomes:

\begin{equation}
\begin{aligned}
\mathcal{L}_{\text{MeanFlow}}(\theta) & = \mathbb{E}_{t, \boldsymbol{x}_t} \left[  \left\|\boldsymbol{F}_\theta\left(\boldsymbol{x}_t, t\right) - \left(
\boldsymbol{v}_t - t\frac{\mathrm{d} \boldsymbol{F}_{\theta^{-}}(\boldsymbol{x}_t, t)}{\mathrm{d}t}\right)\right\|_2^2 \right].
\end{aligned}
\end{equation}

We differentiate the loss with respect to $\theta$. Omitting the constant factor, the gradient becomes:
\begin{equation}
\nabla_\theta \mathcal{L}_{\text{MeanFlow}} (\theta) = \mathbb{E}_{t, \boldsymbol{x}_t} \left[ -\left\langle \nabla_\theta \boldsymbol{F}_\theta(\boldsymbol{x}_t, t), \boldsymbol{v}_t - \boldsymbol{F}_{\theta}(\boldsymbol{x}_t, t) - t \frac{\mathrm{d}\boldsymbol{F}_{\theta^{-}}}{\mathrm{d}t} \right\rangle \right].
\end{equation}

Comparing the resulting gradients:
\begin{equation}
\nabla_\theta \mathcal{L}_{\text{sCM}} = \mathbb{E}_{t, \boldsymbol{x}_t} \left[ -\left\langle t \nabla_\theta \boldsymbol{F}_\theta(\boldsymbol{x}_t, t), \boldsymbol{v}_t - \boldsymbol{F}_{\theta^{-}}(\boldsymbol{x}_t, t) - t \frac{\mathrm{d}\boldsymbol{F}_{\theta^{-}}}{\mathrm{d}t} \right\rangle \right],
\end{equation}

\begin{equation}
\nabla_\theta \mathcal{L}_{\text{MeanFlow}} = \mathbb{E}_{t, \boldsymbol{x}_t} \left[ -\left\langle \nabla_\theta \boldsymbol{F}_\theta(\boldsymbol{x}_t, t), \boldsymbol{v}_t - \boldsymbol{F}_{\theta}(\boldsymbol{x}_t, t) - t \frac{\mathrm{d}\boldsymbol{F}_{\theta^{-}}}{\mathrm{d}t} \right\rangle \right].
\end{equation}

Since $\theta^{-}$ represents the parameters with stopped gradients, numerically $\boldsymbol{F}_{\theta}$ is equal to $\boldsymbol{F}_{\theta^{-}}$. Consequently, the error signal terms (the second term in the inner product) are numerically identical in both methods. The gradients differ only by the weighting factor $t$ present in sCM.

%% file: paper.bbl
\begin{thebibliography}{22}
\providecommand{\natexlab}[1]{#1}
\providecommand{\url}[1]{\texttt{#1}}
\expandafter\ifx\csname urlstyle\endcsname\relax
  \providecommand{\doi}[1]{doi: #1}\else
  \providecommand{\doi}{doi: \begingroup \urlstyle{rm}\Url}\fi

\bibitem[{Black Forest Labs} et~al.(2025){Black Forest Labs}, Batifol, Blattmann, Boesel, Consul, Diagne, Dockhorn, English, English, Esser, Kulal, Lacey, Levi, Li, Lorenz, M{"u}ller, Podell, Rombach, Saini, Sauer, and Smith]{flux1kontext2025}
{Black Forest Labs}, Stephen Batifol, Andreas Blattmann, Frederic Boesel, Saksham Consul, Cyril Diagne, Tim Dockhorn, Jack English, Zion English, Patrick Esser, Sumith Kulal, Kyle Lacey, Yam Levi, Cheng Li, Dominik Lorenz, Jonas M{"u}ller, Dustin Podell, Robin Rombach, Harry Saini, Axel Sauer, and Luke Smith.
\newblock Flux{.}1 kontext: Flow matching for in-context image generation and editing in latent space.
\newblock \emph{arXiv preprint arXiv:2506.15742}, June 2025.

\bibitem[Chen et~al.(2025)Chen, Xue, Zhao, Yu, Paul, Chen, Cai, Han, and Xie]{chen2025sana}
Junsong Chen, Shuchen Xue, Yuyang Zhao, Jincheng Yu, Sayak Paul, Junyu Chen, Han Cai, Song Han, and Enze Xie.
\newblock Sana-sprint: One-step diffusion with continuous-time consistency distillation.
\newblock \emph{arXiv preprint arXiv:2503.09641}, 2025.

\bibitem[Freepik(2024)]{fluxlite}
Freepik.
\newblock Flux.1-lite-8b.
\newblock \url{https://huggingface.co/Freepik/flux.1-lite-8B}, 2024.
\newblock Hugging Face model repository.

\bibitem[Geng et~al.(2025)Geng, Deng, Bai, Kolter, and He]{geng2025mean}
Zhengyang Geng, Mingyang Deng, Xingjian Bai, J.~Zico Kolter, and Kaiming He.
\newblock Mean flows for one-step generative modeling.
\newblock In \emph{NeurIPS}, 2025.

\bibitem[{Google DeepMind}(2025)]{gdm2025imagen4}
{Google DeepMind}.
\newblock Imagen 4 model card.
\newblock Model card (PDF), May 2025.
\newblock \url{https://storage.googleapis.com/deepmind-media/Model-Cards/Imagen-4-Model-Card.pdf}.
\newblock Published May 20, 2025.

\bibitem[Ho et~al.(2020)Ho, Jain, and Abbeel]{ho2020denoising}
Jonathan Ho, Ajay Jain, and Pieter Abbeel.
\newblock Denoising diffusion probabilistic models.
\newblock \emph{NeurIPS}, 33:\penalty0 6840--6851, 2020.

\bibitem[Karras et~al.(2022)Karras, Aittala, Aila, and Laine]{karras2022elucidating}
Tero Karras, Miika Aittala, Timo Aila, and Samuli Laine.
\newblock Elucidating the design space of diffusion-based generative models.
\newblock In \emph{NeurIPS}, 2022.

\bibitem[Lu and Song(2025)]{lu2025simp}
Cheng Lu and Yang Song.
\newblock Simplifying, stabilizing \& scaling continuoustime consistency models.
\newblock In \emph{ICLR}, 2025.

\bibitem[Luhman and Luhman(2021)]{luhman2021knowledge}
Eric Luhman and Troy Luhman.
\newblock Knowledge distillation in iterative generative models for improved sampling speed.
\newblock \emph{arXiv preprint arXiv:2101.02388}, 2021.

\bibitem[Luo et~al.(2023)Luo, Tan, Huang, Li, and Zhao]{luo2023lcm}
Simian Luo, Yiqin Tan, Longbo Huang, Jian Li, and Hang Zhao.
\newblock Latent consistency models: Synthesizing high-resolution images with few-step inference.
\newblock \emph{arXiv preprint arXiv:2310.04378}, 2023.

\bibitem[{ModelTC}(2024)]{qwenimagelig}
{ModelTC}.
\newblock Qwen-image-lightning.
\newblock \url{https://huggingface.co/lightx2v/Qwen-Image-Lightning}, 2024.

\bibitem[Rombach et~al.(2022)Rombach, Blattmann, Lorenz, Esser, and Ommer]{rombach2022high}
Robin Rombach, Andreas Blattmann, Dominik Lorenz, Patrick Esser, and Bj{\"o}rn Ommer.
\newblock High-resolution image synthesis with latent diffusion models.
\newblock In \emph{CVPR}, pages 10684--10695, 2022.

\bibitem[Saharia et~al.(2022)Saharia, Chan, Saxena, Li, Whang, Denton, and et~al.]{saharia2022photorealistic}
Chitwan Saharia, William Chan, Saurabh Saxena, Lala Li, Jay Whang, Emily Denton, and et~al.
\newblock Photorealistic text-to-image diffusion models with deep language understanding.
\newblock \emph{Advances in Neural Information Processing Systems (NeurIPS)}, 2022.

\bibitem[Salimans and Ho(2022)]{salimans2022progressive}
Tim Salimans and Jonathan Ho.
\newblock Progressive distillation for fast sampling of diffusion models.
\newblock \emph{arXiv preprint arXiv:2202.00512}, 2022.

\bibitem[Sauer et~al.(2024)Sauer, Boesel, Dockhorn, Blattmann, Esser, and Rombach]{sauer2024fast}
Axel Sauer, Frederic Boesel, Tim Dockhorn, Andreas Blattmann, Patrick Esser, and Robin Rombach.
\newblock Fast high-resolution image synthesis with latent adversarial diffusion distillation.
\newblock In \emph{SIGGRAPH Asia 2024 Conference Papers}, pages 1--11, 2024.

\bibitem[Sohl{-}Dickstein et~al.(2015)Sohl{-}Dickstein, Weiss, Maheswaranathan, and Ganguli]{sohl2015deep}
Jascha Sohl{-}Dickstein, Eric Weiss, Niru Maheswaranathan, and Surya Ganguli.
\newblock Deep unsupervised learning using nonequilibrium thermodynamics.
\newblock In \emph{ICML}, 2015.

\bibitem[Song et~al.(2020)Song, Meng, and Ermon]{song2020denoising}
Jiaming Song, Chenlin Meng, and Stefano Ermon.
\newblock Denoising diffusion implicit models.
\newblock \emph{arXiv preprint arXiv:2010.02502}, 2020.

\bibitem[Song et~al.(2023)Song, Dhariwal, Chen, and Sutskever]{song2023consistency}
Yang Song, Prafulla Dhariwal, Mark Chen, and Ilya Sutskever.
\newblock Consistency models.
\newblock \emph{arXiv preprint arXiv:2303.01469}, 2023.

\bibitem[Wang et~al.(2023)Wang, Zheng, He, Chen, and Zhou]{wang2022diffusion}
Zhendong Wang, Huangjie Zheng, Pengcheng He, Weizhu Chen, and Mingyuan Zhou.
\newblock Diffusion-gan: Training gans with diffusion.
\newblock In \emph{ICLR}, 2023.

\bibitem[Wu et~al.(2025)Wu, Li, Zhou, Lin, Gao, Yan, Yin, Bai, Xu, Chen, et~al.]{wu2025qwen}
Chenfei Wu, Jiahao Li, Jingren Zhou, Junyang Lin, Kaiyuan Gao, Kun Yan, Sheng-ming Yin, Shuai Bai, Xiao Xu, Yilei Chen, et~al.
\newblock Qwen-image technical report.
\newblock \emph{arXiv preprint arXiv:2508.02324}, 2025.

\bibitem[Zheng et~al.(2025)Zheng, Wang, Ma, Chen, Zhang, Balaji, Chen, Liu, Zhu, and Zhang]{zheng2025large}
Kaiwen Zheng, Yuji Wang, Qianli Ma, Huayu Chen, Jintao Zhang, Yogesh Balaji, Jianfei Chen, Ming-Yu Liu, Jun Zhu, and Qinsheng Zhang.
\newblock Large scale diffusion distillation via score-regularized continuous-time consistency.
\newblock \emph{arXiv preprint arXiv:2510.08431}, 2025.

\bibitem[Zhou et~al.(2025)Zhou, Ermon, and Song]{zhou2025inductive}
Linqi Zhou, Stefano Ermon, and Jiaming Song.
\newblock Inductive moment matching.
\newblock In \emph{ICML}, 2025.

\end{thebibliography}
